%% file: main.tex
\documentclass{article}
\usepackage{amssymb}
\usepackage{graphicx}
\usepackage{amsmath}
\usepackage{amssymb}
\usepackage{tikz}

\usepackage{enumitem}
\usepackage{algorithm}
\usepackage{algpseudocode}
\usepackage{adjustbox}
\usepackage{booktabs}
\usepackage{tcolorbox}
\usepackage{array}
\usepackage{svg}
\usepackage{cite}
\usepackage[subrefformat=parens]{subcaption}
\usepackage[preprint]{corl_2025} 

\title{Few-Shot Neuro-Symbolic Imitation Learning for
Long-Horizon Planning and Acting}

\author{
  Pierrick Lorang\\
  Human-Robot Interaction Lab\\
  Tufts University, 
  United States\\
  \& Austrian Institute of Technology,
  Austria\\
  \texttt{first.last@tufts.edu} \\
  \And
  Hong Lu\\
  Human-Robot Interaction Lab\\
  Tufts University, 
  United States\\
  \texttt{first.last663424@tufts.edu} \\
  \AND
  Johannes Huemer \\
  Austrian Institute of\\ Technology, 
  Austria\\
  \texttt{first.last@ait.ac.at} \\
  \And
  Patrik Zips \\
  Austrian Institute of\\ Technology,
  Austria\\
  \texttt{first.last@ait.ac.at} \\
  \And
  Matthias Scheutz \\
  Human-Robot Interaction Lab\\
  Tufts University, 
  United States\\
  \texttt{first.last@tufts.edu} \\
}

\begin{document}
\maketitle


\begin{abstract}
    Imitation learning enables intelligent systems to acquire complex behaviors with minimal supervision. However, existing methods often focus on short-horizon skills, require large datasets, and struggle to solve long-horizon tasks or generalize across task variations and distribution shifts. We propose a novel neuro-symbolic framework that jointly learns continuous control policies and symbolic domain abstractions from a few skill demonstrations. Our method abstracts high-level task structures into a graph, discovers symbolic rules via an Answer Set Programming solver, and trains low-level controllers using diffusion policy imitation learning. A high-level oracle filters task-relevant information to focus each controller on a minimal observation and action space. Our graph-based neuro-symbolic framework enables capturing complex state transitions, including non-spatial and temporal relations, that data-driven learning or clustering techniques often fail to discover in limited demonstration datasets. 
    We validate our approach in six domains that involve four robotic arms, Stacking, Kitchen, Assembly, and Towers of Hanoi environments, and a distinct Automated Forklift domain with two environments. The results demonstrate high data efficiency with as few as five skill demonstrations, strong zero- and few-shot generalizations, and interpretable decision making. A video of our results is available at \href{https://youtu.be/E8slaN81oAA}{this link}.

\end{abstract}



\input{introduction}
\input{related_work}

\input{preliminaries}
\input{method}

\input{evaluation}

\input{results}


\section{Conclusion}
\label{sec:conclusion}

Our framework advances explainable, generalizable, and data-efficient long-horizon task execution, addressing key challenges in modern AI. By allowing agents to generalize across diverse tasks and environments, including novel configurations and previously unseen problems, it demonstrates high performance in solving complex reasoning tasks with only a few demonstrations. Furthermore, our approach scales efficiently to more complex tasks with minimal additional supervision, making it a powerful paradigm for human-taught robotics. Importantly, it eliminates the need for direct environment interaction during learning, enabling agents to plan and act effectively without risky exploration or reliance on simulated models or digital twins. Thus, we have demonstrated the utility of our approach in robotics and industrial settings, paving the way for more efficient, scalable, and human-friendly solutions in real-world applications.
\input{limitations}
\clearpage


\bibliography{references}  

\clearpage
\input{appendix}

\end{document}

%% file: introduction.tex
\section{Introduction}

\begin{figure}[ht]
  \centering
  \begin{minipage}[]{1.0\textwidth}
    \includegraphics[width=1.0\textwidth]{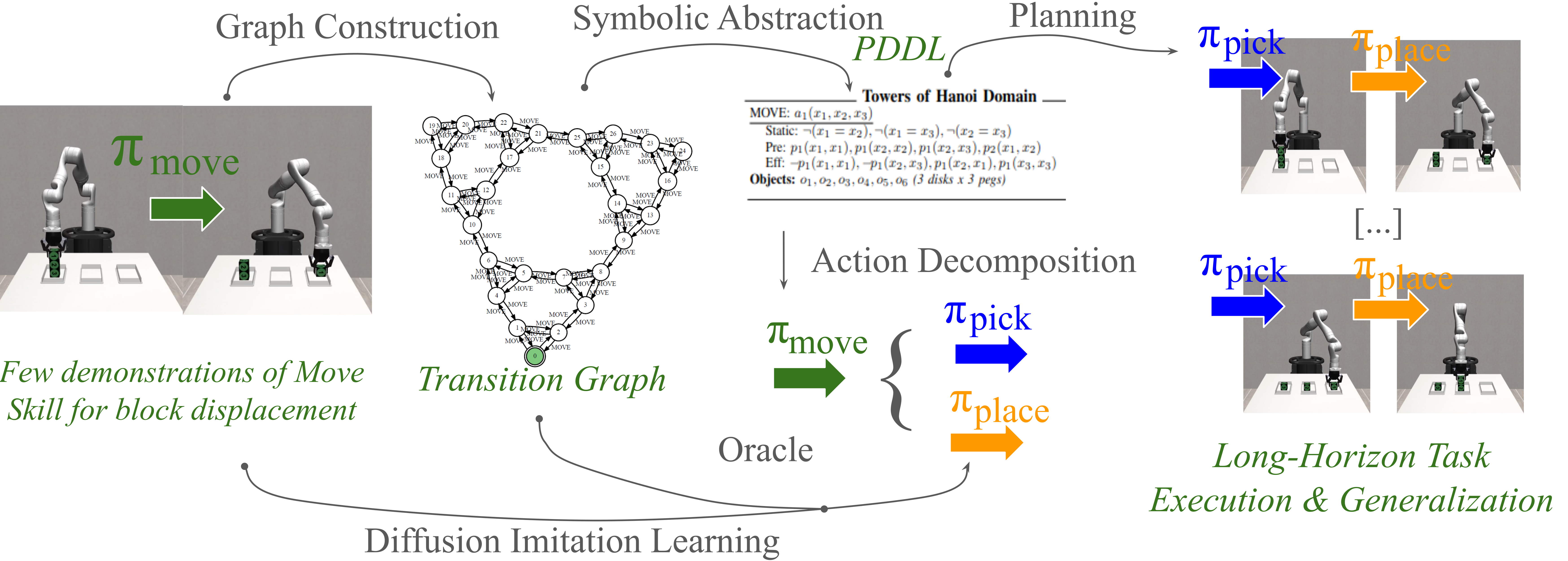}
    \small \caption{Our neuro-symbolic framework integrates graph construction, symbolic abstraction, planning, action decomposition, imitation learning and space filtering. Starting with just a few skill demonstrations (left), we construct a transition graph capturing transitions (edges), which are skill transitions between two high-level states (black-box nodes). This graph enables automatic PDDL model extraction via an ASP solver, which powers high-level planning. The oracle decomposes complex tasks into primitive action step policies executed by learned diffusion-based controllers, allowing generalization to novel long-horizon tasks while requiring minimal training data.}
    \label{fig:inputdata}
    \end{minipage}
\end{figure}

Teaching robots complex tasks remains a central challenge in robotics and artificial intelligence. Imitation learning has emerged as a prominent solution that enables robots to acquire behaviors from demonstrations~\cite{Hussein_Gaber_Elyan_Jayne_2017,Osa_Pajarinen_Neumann_Bagnell_Abbeel_Peters_2018,Fang_Jia_Guo_Xu_Wen_Sun_2019,Zheng_Verma_Zhou_Tsang_Chen_2022,Zare_Kebria_Khosravi_Nahavandi_2024}. However, it typically focuses on short-horizon skills, struggles with distribution shifts over time~\cite{Rajaraman_Yang_Jiao_Ramachandran_2020, Xu_Li_Yu_2020}, and generalizes poorly to novel situations.

These limitations are magnified in long-horizon tasks, where successful behavior demands not just precise low-level control but strategic high-level planning. Humans excel by abstracting problems into symbolic representations, facilitating reasoning and generalization. Hierarchical approaches such as Task and Motion Planning (TAMP)~\cite{Wolfe_Marthi_Russell_2010, Kaelbling_Lozano-Pérez_2011, Garrett_Chitnis_Holladay_Kim_Silver_Kaelbling_Lozano-Pérez_2021} leverage a similar divide, but traditionally rely on manually crafted symbolic models, making them brittle and laborious to adapt.
Previous work has sought to learn either symbolic models~\cite{Arora_Fiorino_Pellier_Métivier_Pesty_2018, Konidaris_Kaelbling_Lozano-Perez_2018, Bonet_Geffner_2020, kr2021rbrg, Silver_Chitnis_Tenenbaum_Kaelbling_Lozano-Perez_2021, Verma_Marpally_Srivastava_2022, Chitnis_Silver_Tenenbaum_Lozano-Perez_Kaelbling_2022, Ahmetoglu_Seker_Piater_Oztop_Ugur_2022, Silver_Chitnis_Kumar_McClinton_Lozano-Perez_Kaelbling_Tenenbaum_2022, Li_Silver_2023, Kumar_McClinton_Chitnis_Silver_Lozano-Pérez_Kaelbling_2023, Iii_B_2024, Shah_Nagpal_Verma_Srivastava_2024, Umili_Antonioni_Riccio_Capobianco_Nardi_Giacomo} or low-level controllers~\cite{Yang_Lyu_Liu_Gustafson_2018, Illanes_Yan_Icarte_McIlraith_2020, Kokel_Manoharan_Natarajan_Ravindran_Tadepalli_2021, Guan_Sreedharan_Kambhampati, Balloch_Lin_Wright_Peng_Hussain_Srinivas_Kim_Riedl_2023, Cheng_Xu_2023, Silver_Athalye_Tenenbaum_Lozano-Pérez_Kaelbling_2023, Acharya_Raza_Dourado_Velasquez_Song_2023, Cheng_Xu_2023, Lorang_Goel_Shukla_Zips_Scheutz_2024, Lorang_Horvath_Kietreiber_Zips_Heitzinger_Scheutz_2024, Goel_Lymperopoulos_Thielstrom_Krause_Feeney_Lorang_Schneider_Wei_Kildebeck_Goss_et_al._2024} independently. Some more recent work proposes learning both layers from demonstrations~\cite{Silver_Athalye_Tenenbaum_Lozano-Pérez_Kaelbling_2023}. However, these methods typically assume access to symbolic abstractions \emph{a priori}: either through manually designed symbolic state representations, known predicates and object types, or predefined mapping functions from continuous states to symbolic facts.


To our knowledge, no prior work jointly learns low-level control policies and high-level planning models from a few demonstrations without relying on predefined symbolic states, predicates, or lexicon. In our approach, each demonstration is modeled as a transition between two high-level environment states, represented as black-box nodes with visual snapshots. Identical states
can be matched by a human based solely on visual comparison, requiring no expertise in symbolic structures or planning. The resulting skills and nodes form a domain structure graph, which is automatically constructed and passed to an ASP solver to yield a PDDL-form symbolic domain. This symbolic representation supports learning data-efficient controls, as well as planning during execution.


Our approach can model object relationships beyond static (position based) spatial predicates, as typically done by clustering approaches~\cite{Shah_Nagpal_Verma_Srivastava_2024}, and allows for instance time-dependent transitions abstraction—such as waiting for food to cook—by representing them as temporal edges in the graph.
A symbolic abstraction is then automatically extracted by using an Answer Set Programming (ASP) solver~\cite{Bonet_Geffner_2020, kr2021rbrg} to discover a Planning Domain Definition Language (PDDL)~\cite{mcdermott_pddl_1998} model consistent with the observed transitions.


\textbf{Contributions.} We present the first \emph{neuro-symbolic imitation learning} framework that jointly learns low-level control policies and high-level symbolic abstractions from few raw demonstrations, without predefined states, predicates, lexicons, or domain knowledge. From these abstractions, an oracle automatically segments demonstrations, filters irrelevant information, and trains neural controllers for each operator. Our method provides: (1) data-efficient learning from few demonstrations, (2) robust generalization to out-of-distribution tasks, and (3) scalable continual learning. We validate it across six domains—Stacking, Kitchen, Assembly, Towers of Hanoi, and two forklift tasks—demonstrating strong data efficiency, broad applicability, robust generalization, and interpretable symbolic plans. 


%% file: related_work.tex
\section{Related Work}

Prior work on bi-level architectures often assumes access to symbolic representations. Some approaches learn action models from symbolic traces~\cite{Konidaris_Kaelbling_Lozano-Perez_2018, chitnis2021learning, Umili_Antonioni_Riccio_Capobianco_Nardi_Giacomo, Chitnis_Silver_Tenenbaum_Lozano-Perez_Kaelbling_2022, Kumar_McClinton_Chitnis_Silver_Lozano-Pérez_Kaelbling_2023}, relying on predefined grammars and predicates. Others guide low-level RL with symbolic domains~\cite{Icarte2020RewardME, sarathy2021spotter, goel2022rapidlearn, Gehring_Asai_Chitnis_Silver_Kaelbling_Sohrabi_Katz_2022, Cheng_Xu_2023, peorl-Yang, Lorang_Goel_Shukla_Zips_Scheutz_2024, Lorang_Lu_Scheutz_2025}, but still require manually crafted symbolic states.

Learning skill sequences from demonstrations~\cite{Manschitz_Kober_Gienger_Peters_2014, Le_Jiang_Agarwal_Dudik_Yue_Hal_Daumé_2018, Pertsch_Lee_Wu_Lim_2021, Tanwani_Yan_Lee_Calinon_Goldberg_2021, Zhu_Stone_Zhu_2022, Teng_Chen_Ai_Zhou_Xuanyuan_Hu_2023} enables long-horizon imitation but lacks modularity, explainability, and data efficiency compared to symbolic planning. Recent approaches extract symbolic abstractions from raw data by clustering to induce predicates~\cite{Shah_Nagpal_Verma_Srivastava_2024, Keller_Tanneberg_Peters_2025}; however, such approach still require fifty or more demonstrations.
Segmentation~\cite{Loula_Allen_Silver_Tenenbaum_2020} helps but cannot fully remove this reliance. SAT- or ASP-based model learning~\cite{Bonet_Geffner_2020, kr2021rbrg} reduces data requirements for symbolic domain acquisition. Prior work typically uses these solvers to learn continuous features that improve TAMP efficiency from a few example plans with embedded continuous data~\cite{Curtis_Silver_Tenenbaum_Lozano-Pérez_Kaelbling_2022}. However, these approaches assume a predefined symbolic domain and solved plans, whereas we jointly learn both the symbolic domain and continuous controllers directly from a handful of raw demonstrations.

In this work, we combine the abstraction of the planning domain based on ASP together with continuous controllers learning from few skill demonstrations, enabling scalable long-horizon task solving with minimal supervision.

%% file: preliminaries.tex
\section{Preliminaries}

\textbf{Symbolic Planning.} Symbolic planning builds upon a formal domain description $\sigma = \langle \mathcal{E}, \mathcal{F}, \mathcal{S}, \mathcal{O}\rangle$, where $\mathcal{E}$ is a set of entities, $\mathcal{F}$ a set of boolean or numerical predicates over entities, $\mathcal{S}$ a set of symbolic states formed by grounded predicates, and $\mathcal{O}$ a set of operators. Each operator $o \in \mathcal{O}$ is defined by preconditions $\psi$ and effects $\omega$ over predicates. A grounded operator $\hat{o}$ binds objects to parameters and can be applied if its preconditions hold, updating the state according to its effects. A planning task $T=(\mathcal{E}, \mathcal{F}, \mathcal{O}, s_0, s_g)$ seeks a plan $\mathcal{P}=[o_1,\ldots,o_{|\mathcal{P}|}]$ that transitions from initial state $s_0$ to goal state $s_g$~\cite{mcdermott_pddl_1998}.

\textbf{Imitation Learning (IL).} IL aims to learn a policy $\pi(\tilde{s})$ from expert demonstrations 
$\{(\tilde{s}_t, a_t, \tilde{s}_{t+1})\}_{t=0}^{T}$, 
where $\tilde{s}_t$ is a continuous state, $a_t$ the expert action, and $\tilde{s}_{t+1}$ the resulting state. 
The policy minimizes the mean squared error between predicted and expert actions, defined as
$L(\pi) = \frac{1}{T} \sum_{t=0}^{T} \left\| \pi(\tilde{s}_t) - a_t \right\|^2$. Unlike reinforcement learning, IL avoids exploration and reward engineering, enabling more data-efficient learning of complex behaviors from demonstrations.

\textbf{Neuro-Symbolic Architecture.} Neuro-symbolic architectures combine symbolic reasoning with neural control. A planner solves a STRIPS task $T = \langle \mathcal{E}, \mathcal{F}, \mathcal{O}, s_0, s_g \rangle$ to produce a plan $\mathcal{P}=[o_1,\ldots,o_{|\mathcal{P}|}]$, where each operator $o_i$ is refined into a neural skill $\pi_i \in \Pi$. Each skill $\pi_i$ interacts with the environment to realize the operator's effects $\omega_i$, transitioning the system from a state $s$ to a new state $s'$. This layered approach enables flexible execution in continuous spaces while maintaining high-level task abstraction.



\textbf{Problem Formulation.} 
We consider a dataset of object-space skill demonstrations 
$\mathcal{D} = \{\tilde{\tau}_0, \dots, \tilde{\tau}_{|\mathcal{D}|}\}$, 
where each trajectory $\tilde{\tau}_i = \{(\tilde{s}_t, a_t, \tilde{s}_{t+1})\}_{t=0}^{|\tilde{\tau}_i|-1}$ 
captures state-action transitions over objects $\varepsilon \in \mathcal{E}$. 
Each demonstration also provides two images: one of the initial state $v$ and one of the final state $v'$.  
Human input is limited to two forms:
(1) demonstrate individual lifted skills, with skill labels shared across objects 
(2) match high-level states through image comparison. No symbolic annotations, object types, or semantics are assumed.

%% file: method.tex
\section{Neuro-Symbolic Imitation Learning}



After receiving the set of raw demonstration trajectories \( \mathcal{D}\), our neuro-symbolic agent aims to learn robust, generalizable solutions for long-horizon Task and Motion Planning (TAMP) problems. Each trajectory is mapped
to a node transition \( \mathcal{\tau}_i^{node} = \zeta(\tilde{\tau}_i) = (n_{\text{start}}, l, n_{\text{end}})\) where $n_\text{start}$ and $n_\text{end}$ denote abstract high-level states at the beginning and end of the skill demonstration, and $l$ is a human-assigned label describing the transition between them. This process structures the skill demonstrations into a task-space abstraction (Fig.~\ref{fig:inputdata}). The nodes correspond to abstract high-level states (whose symbolic representation remains undiscovered), and the edges represent transitions induced by action operators. A skill typically transitions the environment between two high level states, i.e., performs a node transition, providing a task-space abstraction for the agent. Such skill abstraction can be represented as an edge $(n, l, n')$ in a graph $G$, while $n$ and $n'$ can be paired to their respective visual snapshots $v$ and $v'$, capturing the beginning and the end of the skill.

To leverage this structure, we adopt a bi-level neuro-symbolic learning approach combining imitation learning and symbolic planning. At the symbolic level, the agent constructs a graph from node transitions to infer operators \( \mathcal{O} \) and predicates \( \mathcal{F} \) that abstract the task-space transitions. At the skill level, the agent learns neural policies \( \pi_i \in \Pi \) that realize each operator \( o_i \in \mathcal{O} \) by imitating corresponding segments of the demonstration trajectories.

\begin{figure}[!t]
  \centering
  \begin{minipage}[]{1.0\textwidth}
      \includegraphics[width=0.90\textwidth, height=0.18\textwidth]{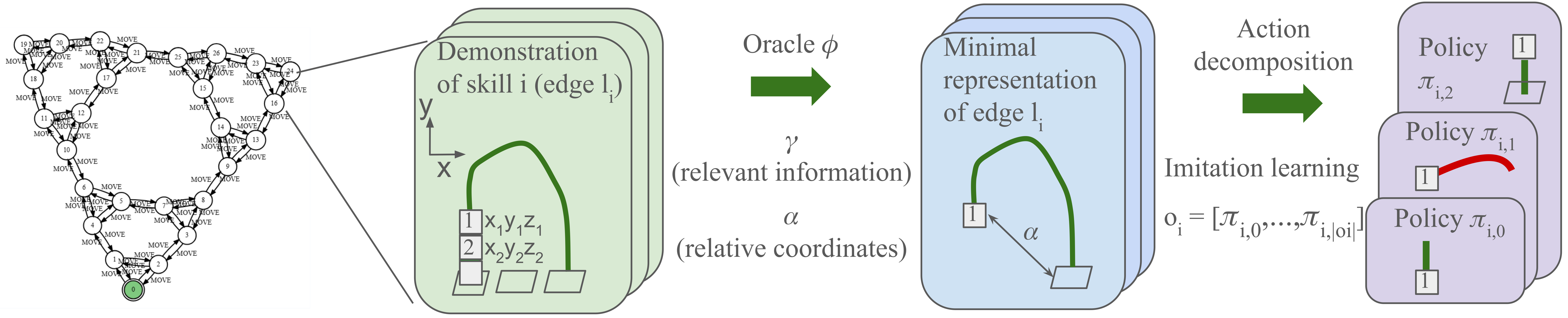}
      \caption{Shown here is an example demonstration of the MOVE operator in the Towers of Hanoi domain, where block 1 is moved off block 2 and placed onto a platform. The agent partitions skill demonstrations into action steps, with an oracle $\phi$ filtering observations to simplify learning. The demonstration is first collected, then filtered using $\gamma$ to retain operator-relevant objects (block 1 and platform 3) and $\alpha$ to express coordinates relative to the end-effector. The trajectory is then decomposed into a sequence of simpler action steps. This enables efficient training of low-level controllers, which are sequenced to execute each symbolic operator~\textit{(Alg.~\ref{alg:Execution}.line 5)}.}
      \label{fig:partition}
      \end{minipage}
\end{figure}

Inspired by the options framework in Hierarchical Reinforcement Learning~\cite{SUTTON1999181}, we further decompose each skill into sequential action steps. Action steps are automatically identified through consistent sequential action space patterns.
For instance, a \textit{MOVE} operator in a \textit{Pick \& Place} task decomposes into \textit{reach pick}, \textit{pick}, \textit{reach drop}, and \textit{drop} stages, each constrained to simpler action spaces (e.g., end-effector position or gripper aperture). This decomposition reduces the complexity of individual policies and restricts their action spaces, simplifying learning (see Fig.~\ref{fig:partition}, and~\cite{SUTTON1999181}).

After acquiring the symbolic operators and their associated neural skills, planning proceeds by specifying an initial and a goal node translated within a PDDL planning problem. A classical planner, MetricFF~\cite{hoffmann2003metric}, computes an abstract plan \( \mathcal{P} = [o_1, \dots, o_{|\mathcal{P}|}] \) mapping operators to their corresponding neural skills~\textit{(Alg.~\ref{alg:Execution}.line 10)}. Execution unfolds by sequentially invoking the associated policies \( \pi_i \), each internally organized into action-step sub-policies \( \pi_{i,j} \) executed until a learned termination condition, modeled by a learned function approximation \( \pi_{term} \), is met~\textit{(Alg.~\ref{alg:Execution}.lines 11-17)}.

This hierarchical and modular framework enables the agent to generalize beyond the demonstrations to unseen tasks, adapt to different object configurations, and robustly solve complex, long-horizon problems with limited training data.

\subsection{Learning Symbolic Structures from Sparse Demonstrations}
\label{sec:learning_graph_sat}

Our method constructs a symbolic graph from unordered demonstrations with minimal human input~\textit{(Alg.~\ref{alg:Execution}.line 1)}. When the agent reaches a new high-level state—known at that moment as a black-box node \( n' \) distinct from the current state \( n \)—the human (1) assigns a transition label \( l \), and (2) links \( n' \) to existing nodes by matching visual snapshots $v$ and $v'$ paired with each node. This process adds an edge \( (n, l, n') \) to the evolving graph \( G = \langle V, E, L \rangle \).

After collecting demonstrations, we compute a minimal bisimulation $\bar{G}$ of $G$ (Alg.~\ref{alg:Execution}, line~2) to eliminate redundant structure (Figs.~\ref{fig:fig1}--\ref{fig:fig4}). This keeps the graph compact, reducing both the search space for domain learning and the effort required for annotation. Formally, for two labeled graphs \( G_1 = \langle V_1, E_1, L_1 \rangle \) and \( G_2 = \langle V_2, E_2, L_2 \rangle \), a bisimulation relation \( R \subseteq V_1 \times V_2 \) satisfies: If \( (s_1, s_1') \in E_1 \) labeled by \( l \), then there exists \( (s_2, s_2') \in E_2 \) with the same label and \( (s_1', s_2') \in R \), and vice versa.
If such an \( R \) exists, \( G_1 \) and \( G_2 \) are bisimilar.

To extract first-order symbolic representations from the graph, we adopt the ASP-based framework of~\cite{Bonet_Geffner_2020, kr2021rbrg}~\textit{(Alg.~\ref{alg:Execution}.line 3)}. The goal is to find the simplest planning instance \( P = \langle \sigma, I \rangle \), where \( \sigma \) is the domain theory and \( I = \langle \mathcal{E}, \text{Init}, \text{Goal} \rangle \) defines instance-specific objects and grounded states. Although \( \text{Init} \) and \( \text{Goal} \) need \textbf{not} be logically specified during demonstrations, they serve to maintain consistency within the symbolic notation.

Each \( P \) defines a labeled graph \( G(P) \) where nodes correspond to symbolic states and labeled edges represent action transitions. We solve for \( \sigma \) such that \( G(P) \) is isomorphic to the input graph \( G \), extending naturally to multiple demonstrations \( G_1, \dots, G_k \) by learning a shared domain \( \sigma \) and separate instances \( P_i = \langle \sigma, I_i \rangle \). We assume graphs are complete and noise-free, and that action labels provide no structural or predicate-level information. 
Finally, the learned domain \( \sigma \) is expressed in PDDL, enabling classical planners to operate over the abstracted symbolic space.

\begin{algorithm}[t]
\footnotesize
\caption{~\textit{\textbf{Neuro-Symbolic Imitation Learning}~($\mathcal{D}, \zeta, \mathcal{E}$)~}}\label{alg:Execution}
\begin{algorithmic}[1]
\Require A set of raw demonstration trajectories $\mathcal{D} = \{\tilde{\tau}_0, \dots, \tilde{\tau}_{|\mathcal{D}|}\}$
\Require An automatic function $\zeta$ that extracts \( \mathcal{\tau}^{node} = \zeta(\tilde{\tau}) \) 
\Require A set of entities $\varepsilon \in \mathcal{E}$

\Statex \textbf{\underline{Learning Phase}}

\State $\mathcal{G} \leftarrow$ \textbf{Build\_Graph}($\mathcal{D}, \zeta$) \Comment{Querying Human input, sec.\ref{sec:learning_graph_sat}}
\State $\bar{\mathcal{G}} \leftarrow$ \textbf{Get\_Minimal\_Graph}($\mathcal{G}$) \Comment{Verify if a Bisimulation exists}
\State $\{ \mathcal{F}, \mathcal{O}\} \leftarrow $ \textbf{Abstract}($\bar{\mathcal{G}}$) \Comment{ASP solver, sec.\ref{sec:learning_graph_sat}}
\State $\Pi \leftarrow \{\pi_i \text{, $\pi_i$ is mapped to $o_i$ } \forall o_i \in \mathcal{O}\} $
\State $\forall \pi_i \in \Pi, \{\pi_{i,j}, \mathcal{D}_{o_i, j} \}\leftarrow $ \textbf{Cluster\_Action\_Steps}($\pi_i, \mathcal{D}_{o_i}$) \Comment{Fig.~\ref{fig:partition}, Split Skills into Steps}
\State $\forall \pi_{i,j} \in \Pi,$ \textbf{ Obs}($\pi_{i,j}$) $\leftarrow $\textbf{Oracle}($\pi_{i,j}$) \Comment{Enforces $\phi(\tilde{s})$, sec.\ref{sec:oracle}}
\State $\forall \pi_{i,j} \in \Pi$, \textbf{Train}($\pi_{i,j}$, $\mathcal{D}_{o_i,j}$) \Comment{Diffusion policies, sec.\ref{sec:diffusion}}

\Statex \textbf{\underline{Execution Phase}}

\State $\{n_0, n_g\} \leftarrow $ \textbf{QueryTask}() \Comment{Query to select start and goal nodes (via paired state picture)}
\State \( T = \langle \mathcal{E}, \mathcal{F}, \mathcal{O}, s_0, s_g \rangle \)
\State {$\mathcal{P} \leftarrow$ \textbf{Plan}($T$)}\Comment{$\mathcal{P}=\langle o_1, o_2, ..., o_{|\mathcal{P}|}\rangle$}
\For {$o_i \in \mathcal{P}$}
\State $\pi_i \leftarrow $ \textit{mapped}($o_i$) \Comment{Sequential execution of Skills, sec.\ref{sec:neurosym_il}.Intro}
\For {$\pi_{i,j} \in \pi_i$}
\State $\pi_{exec}, \pi_{term} \leftarrow \pi_{i,j}$
\State \textbf{Execute}($\pi_{exec}, \pi_{term}$) \Comment{Sequential execution of Skill Steps}
\EndFor
\EndFor
\end{algorithmic}
\end{algorithm}

\subsection{The Oracle - Filtering Skill-Relevant Data}
\label{sec:oracle}
For each skill, symbolic abstractions identify the critical objects and relations needed for state transitions, ensuring that skills receive only the observations relevant to their symbolic operator models~\textit{(Alg.~\ref{alg:Execution}.line 6)}.
Formally, let \( o_i \in \mathcal{O} \) be an operator associated with the skill \( \pi_i \). The operator \( o_i \) defines a symbolic transition between states \( s \) and \( s' \) such that: $s' = o_i(s)$,
where the transition is characterized by changes in a subset of object states \( \mathcal{E}_i \subseteq \mathcal{E} \), where \( \mathcal{E} \) is the full set of objects in the environment. Let us call : $
    \mathcal{E}_{o_i} = \{ \varepsilon_k \in \mathcal{E} \mid \text{symbolic state of } \varepsilon_k \text{ changes under } o_i \},
$ 
the subset of objects relevant to the grounded operator \( o_i \in \mathcal{O} \).
Consequently, for each skill \( \pi_i \), 
we define the filtered observation function \( \gamma \) that maps the full state \( \tilde{s} \) to a reduced observation space containing only relevant objects:
\begin{equation}
    \gamma(\tilde{s}, o_i) = \tilde{s}({\mathcal{E}}_{o_i}).
\end{equation}

This filtering mechanism ensures that only the relevant objects are considered during skill execution. 

To further enhance efficiency and scalability, we define a transformation function \( \alpha \) that maps absolute object coordinates to coordinates relative to the agent’s end effector.  Then the function $\alpha$ can be expressed as:
$\alpha(\tilde{s}, \mathcal{E}_{o_i}) = \{ \tilde{s}(\varepsilon_k) - \tilde{s}(\text{EE}) \mid \varepsilon_k \in \mathcal{E}_{o_i} \}$,
where \( \tilde{s}(\varepsilon_k) \) represents the absolute coordinates of object \( \varepsilon_k \), and \( \tilde{s}_t(\text{EE}) \) represents the position of the agent’s end effector, i.e., the part that interacts with the objects. This transformation ensures that policies remain invariant to global positioning, improving generalization across different spatial configurations.

We call the function \( \phi \), which applies the object filtering \( \gamma \) and transformation \( \alpha \), \textit{the Oracle}:
\begin{equation}
    \phi(\tilde{s}_t) = \alpha \circ \gamma (\tilde{s}_t, o_i).
\end{equation}

For task execution, a planner generates a symbolic plan 
\(\mathcal{P} = [o_1, o_2, \dots, o_n]\), 
where each operator \(o_i\) is grounded to a specific subset of objects 
\(\mathcal{E}_{o_i}\) in the environment. 
Each operator is then executed using a policy \(\pi_{exec}\), 
whose observation space is determined by \textit{the Oracle} \(\phi\):
\[
    a_t = \pi_{exec}(\phi(\tilde{s}_t)).
\]

For example, if the planner grounds a unary operator \texttt{pick(.)} to the entity \texttt{cube1}, resulting in the grounded operator \(o = \texttt{pick(cube1)}\), the filtering mechanism ensures that the execution policy observes only cube1's relevant properties (e.g., position, orientation, grasp affordances) relative to the agent's proprioception.

The oracle function \( \phi \) is automatically derived from abstractions and requires no human input. It generates a structured perception which reduces observation dimensionality while preserving task-relevant information, improving learning efficiency and scalability.  
Observation filtering thus serves as a symbolic attention mechanism, dynamically focusing observations on essential information at each execution step.

\label{sec:neurosym_il}

\subsection{Learning Continuous Control Policies}
\label{sec:diffusion}

We learn continuous-space control policies from demonstration data \( \mathcal{D} \), where each trajectory \( \tilde{\tau}_i \) consists of state-action pairs \( (\tilde{s}_t, a_t) \)~\textit{(Alg.~\ref{alg:Execution}.line 7)}.  
The objective is to find a policy \( \pi \) minimizing the action prediction loss over demonstrations:
\[
    \pi^* = \arg\min_{\pi \in \Pi} \sum_{\tilde{\tau}_i \in \mathcal{D}} \sum_{t=0}^{T_i} \mathcal{L}(\pi(\tilde{s}_t), a_t),
\]
where \( \mathcal{L} \) measures action prediction error.

To improve generalization, we preprocess states by filtering observations \( \mathcal{E}_{o_i} \) and applying a transformation \( \phi \) to express object positions relative to relevant frames: $\pi^*(\tilde{s}_t) = \pi^*(\phi(\tilde{s}_t))$.

For robustness and sample efficiency, we adopt diffusion policies~\cite{chi2023diffusionpolicy}.
A diffusion model \( p_{\theta}(a_t | \phi(\tilde{s}_t)) \) learns to generate actions by denoising perturbed expert actions, where \( \epsilon_{\theta} \) predicts the added noise: $
    \mathcal{L}_{\text{diff}} = \mathbb{E}_{(\tilde{s}_t, a_t) \sim \mathcal{D}, \epsilon \sim \mathcal{N}(0, I)} [\|\epsilon - \epsilon_{\theta}(a_t + \sigma \epsilon, \phi(\tilde{s}_t))\|^2]$.
Diffusion policies capture multi-modal action distributions, avoiding the mode collapse common in direct regression. Our framework remains compatible with any imitation learning algorithm for control. Actions are hierarchically structured into control spaces (e.g., end-effector motion, gripper control), enabling flexible and modular execution across tasks.

%% file: evaluation.tex
\section{Evaluation}

\begin{figure*}[!t]
    \centering
    \begin{subcaptionbox}{Robosuite: \textit{Stacking} \& \textit{Hanoi}.\label{fig:hanoi}}%
        [0.335\linewidth]
        {\includegraphics[width=\linewidth]{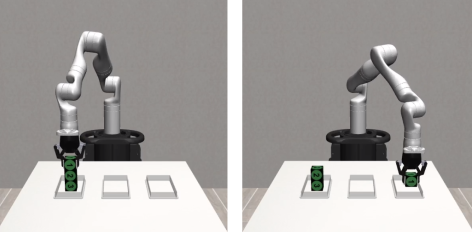}}
    \end{subcaptionbox}
    \hfill
    \begin{subcaptionbox}{Robosuite: \textit{Kitchen}.\label{fig:forklift}}%
        [0.265\linewidth]
        {\includegraphics[width=\linewidth]{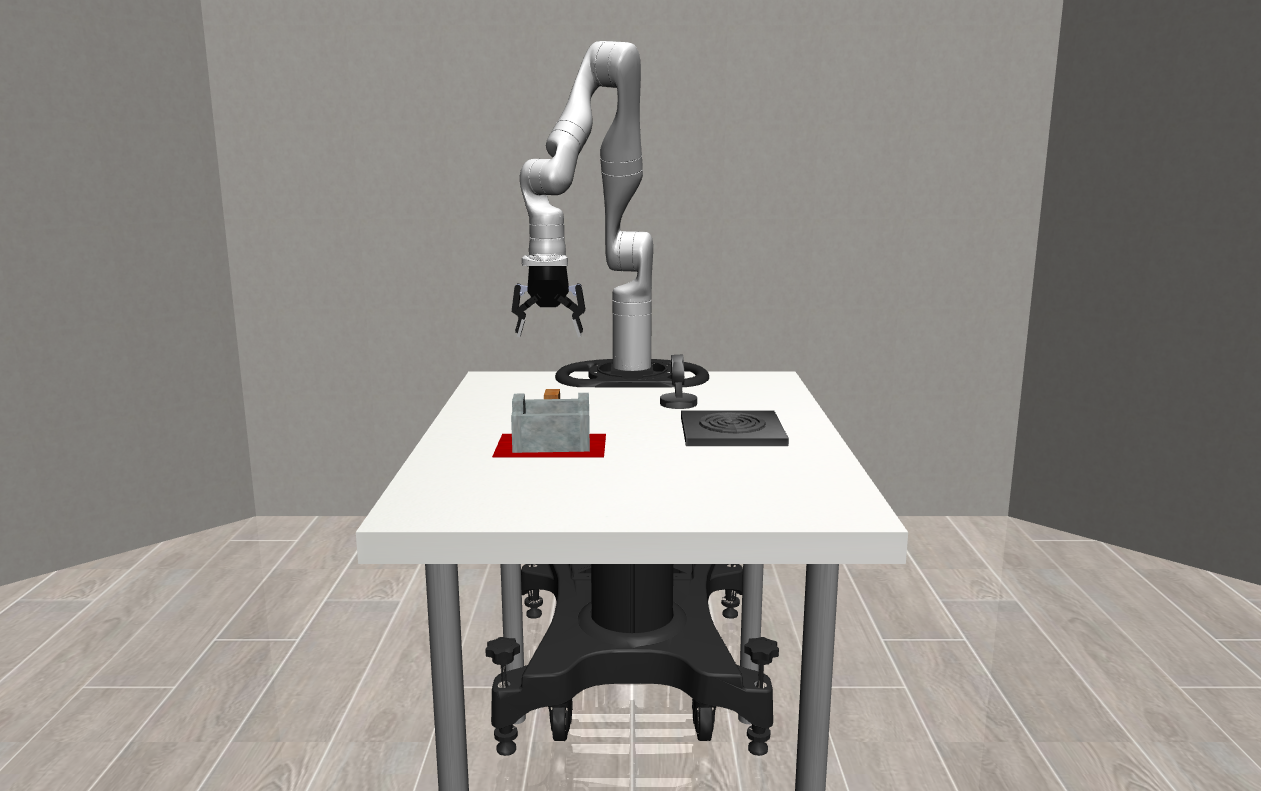}}
    \end{subcaptionbox}
    \hfill
    \begin{subcaptionbox}{ROS2/Gazebo: \textit{Pallets Storage}.\label{fig:kitchenenv}}%
        [0.325\linewidth]
        {\includegraphics[width=\linewidth]{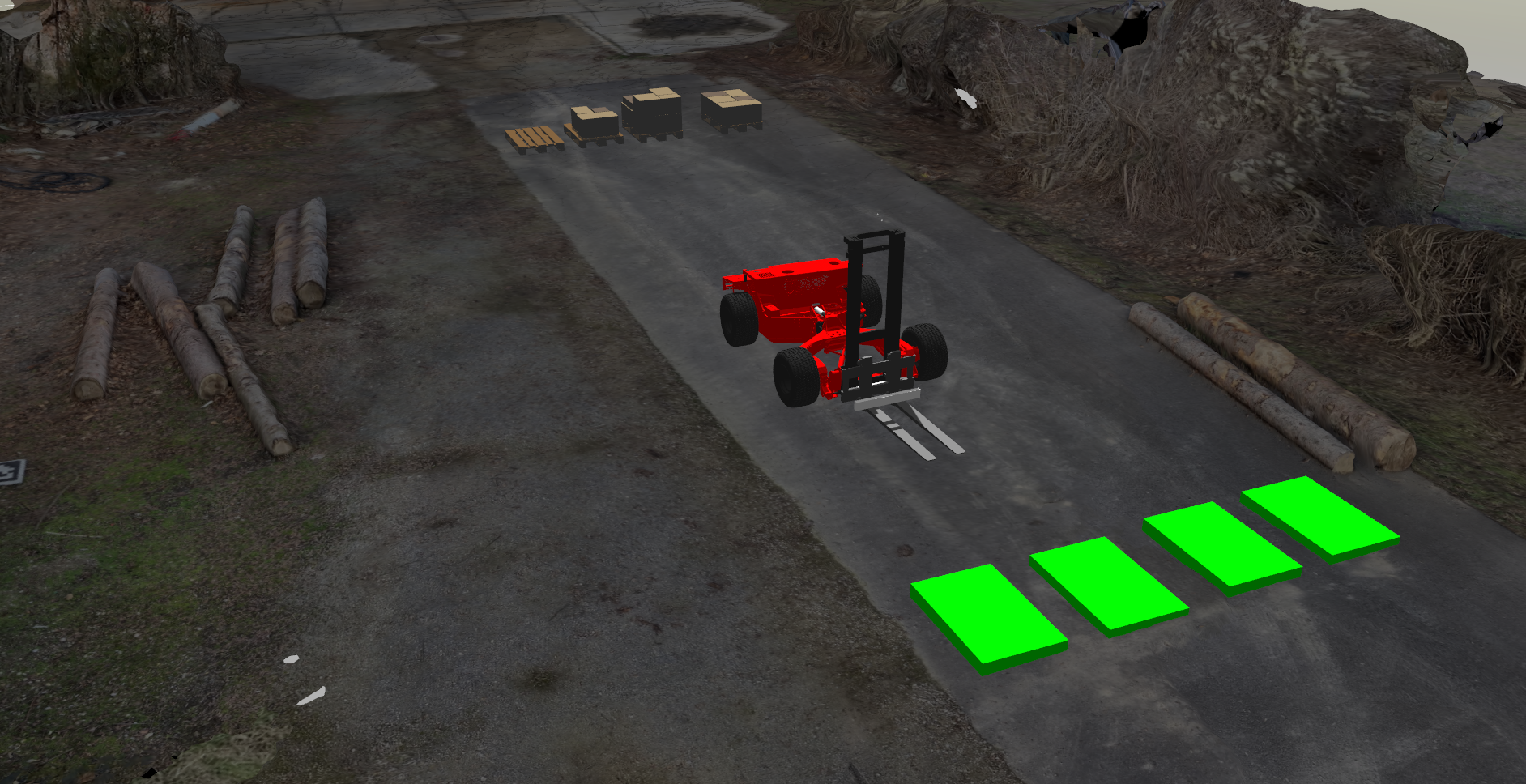}}
    \end{subcaptionbox}
    \caption{Illustrations of some of the simulation domains used for evaluation.}
    \label{fig:comparison}
\end{figure*}

We evaluate our approach across six environments: four in Robosuite including \textit{Stacking} (involving 3 cubes and randomly generated stacking tasks), \textit{Kitchen}, \textit{Nut Assembly} and our own implementation of the \textit{Towers of Hanoi} problem using numerated cubes, as well as completely different environments in ROS 2 \& Gazebo, to perform \textit{Forklift Loading/Unloading tasks)} and \textit{Multiple Pallets Storage tasks} where the agent needs to store multiple pallets at different locations (see Fig.~\ref{fig:comparison}). The forklift’s articulated kinematics in the latter environments introduce significant control challenges, particularly in fork insertion, as the forward displacement is controlled by the rear of the crawler.
Robosuite environments use Cartesian control of the gripper; The ROS 2 \& Gazebo Forklift domain uses motion and forks control. The observations consists of the 6D pose of the objects in the scene and the end-effector (i.e., the part that interacts with the objects). The tasks are randomized during evaluation. The demonstrations provided to our framework are short-horizon skills, for instance stacking, fork insertion, pallet loading, unloading. Each demonstration has a maximal length of 300 steps, and is blurred with noise in the observation dimensions.  

Each agent is evaluated over 30 episodes, and the results are averaged across 5 seeds. Each policy is trained for 8,000 epochs.
\textbf{Metrics}: we use long-horizon tasks success rate for the main experiments. We also display advancement towards completion as evaluation metrics for the generalization experiments.
\textbf{Baselines}: we compare our approach (N-S) against three end-to-end neural baselines:  (1) \textbf{(IL)} Imitation Learning over entire trajectories, (2) \textbf{(H-IL)} Hierarchical Imitation Learning with a high-level policy grounding low-level policies, similar to our oracle reasoning system (3) \textbf{(H-IL Dense)} A similar Hierarchical Imitation Learning baseline, but with the high-level policy receiving the full trajectory instead of a single point at the beginning of each stacking operation. To ensure fair access to information and adherence to the Markov property, the IL baseline policies and the high-level policies of the H-IL baselines receive absolute rather than relative observations. The low-level policies of the H-IL baselines receive relative observations, as transformed by $\alpha$ in our approach.

\textbf{Generalization and Fine Tuning}


\textbf{Zero-shot generalization}:  
We evaluate how well our framework scales to harder Hanoi configurations (4cubes×3pegs to 7cubes×5pegs) and spatial shifts (peg displacements up to 10cm) using only the demonstrations on the most basic configuration (3x3)
In the Forklift domain, we extend the evaluation to \textit{Multiple Pallets Storage tasks} with multiple pallets and zones, also by performing zero-shot transfer.
\textbf{Few-shot fine-tuning}:  
We start from 5 full \textit{MOVE} skill demos, which we incrementally extend over harder Hanoi configurations with 5 additional expert \textit{reach-place} action step demonstrations.
We demonstrate that our agent benefits from continually training on new, more complex, demonstration in a curriculum manner (see Fig.~\ref{fig:generalization_results}).

%% file: results.tex
\section{Results}

\begin{figure}[ht]
  \centering
  \begin{minipage}[]{1.0\textwidth}
    \includegraphics[width=1.0\textwidth]{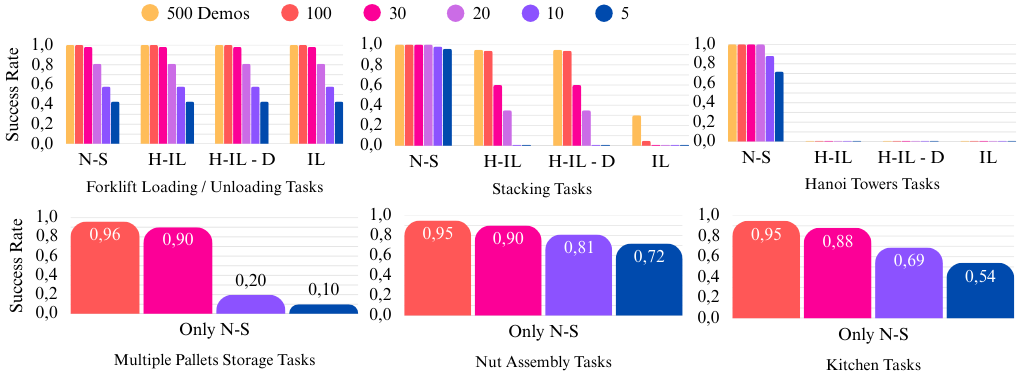}
    \small \caption{Performance comparison between our \textit{Neuro-Symbolic (N-S)} framework and baseline methods. Our approach achieves high success rates on short—\textit{Stacking} \& \textit{Forklift Pallet Loading/Unloading}—and long-horizon tasks—including \textit{Towers of Hanoi}, \textit{Multiple Pallets Storage}, \textit{Nut Assemly} \& \textit{Kitchen}—even with as few as 5 demonstrations. Our approach is domain agnostic, and works in very different scenarios.}
    \label{fig:main_results}
    \end{minipage}
\end{figure}

\textbf{Neural Skills --}  
We isolate neural skill learning in the \textit{Forklift Loading/Unloading} and \textit{Stacking} tasks (Fig.~\ref{fig:main_results} \textit{Above Left \& Middle}), which emphasize short-horizon control rather than planning. All agents perform similarly in the forklift task due to the absence of planning, a single object with a fixed relative pose, and no skill decomposition. In contrast, our agent begins to outperform baselines in \textit{Stacking} by leveraging the \textit{Oracle} for policy refinement, combining relative observations (\( \alpha \)) with symbolic abstractions (\( \gamma \))—a capability baselines lack.
Action space clustering further optimizes skill execution, achieving near 100\% success with as few as 5 demonstrations. The lower success rate in the \textit{Forklift Loading/Unloading} domain stems from the vehicle's articulated kinematics: rear-wheel steering complicates precise fork insertion, and learning these dynamics is challenging in the fork's reference frame. Nonetheless, $30$ demonstrations are sufficient to overcome these difficulties.

\begin{figure}[ht]
  \centering
  \begin{minipage}[]{1.0\textwidth}
    \includegraphics[width=1.0\textwidth]{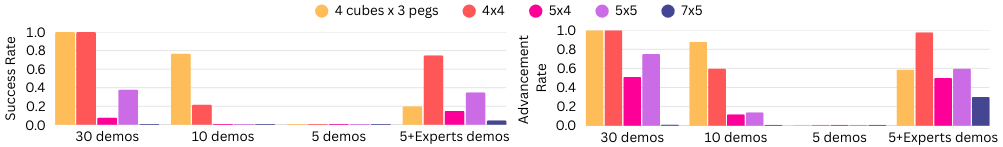}
    \small \caption{Zero- and few-shot generalization results on Different Hanoi Towers configurations.}
    \label{fig:generalization_results}
    \end{minipage}
\end{figure}

\textbf{Reasoning Abilities --}
Reasoning abilities are evaluated in the long-horizon environments (Figs.~\ref{fig:main_results},~\ref{fig:generalization_results}). Baselines completely fail to solve the long-horizon \textit{Towers of Hanoi} task even with 500 full demonstrations, unable to ground sub-goals and diverging into random outputs. In contrast, our method consistently abstracts problem constraints from partial demonstrations, building correct symbolic domains and solving the task reliably. Successes in \textit{Nut Assembly}, \textit{Kitchen} and \textit{Multiple Pallets Storage} further confirms our approach’s domain-independent reasoning. Failures occur only when low-level skills fail to achieve expected effects, not due to reasoning errors.

\textbf{Generalization \& Fine-Tuning --}  
Zero- and few-shot evaluations (Fig.~\ref{fig:generalization_results}) indicate encouraging generalization to new symbolic domains and spatial shifts. With only 5 demonstrations, performance in the 3×3 setting was limited, but adding a few expert corrections (\textit{5+Experts}) yielded significant gains—outperforming agents trained from scratch with 30 demonstrations while using less data overall. In the more complex 7×5 setting, our agent occasionally demonstrated robust behavior, such as recovering from disturbances without replanning, despite occasional control inaccuracies. These results suggest our neuro-symbolic imitation learning framework generalizes well from limited data and benefits from curriculum learning. Importantly, our approach also addresses the two desiderata identified by Silver et al.~\cite{Silver_Athalye_Tenenbaum_Lozano-Pérez_Kaelbling_2023}, in known and unknown settings: (1) our subgoal-conditioned policies and samplers enable reaching diverse low-level states that fulfill the same abstract goal, supporting flexible execution (KD1), and (2) our bilevel symbolic planner can fall back to alternative skill sequences when a plan fails, supporting replanning under abstraction-induced constraints (KD2).

%% file: limitations.tex
\subsection{Limitations and Future Work} 
\label{sec:limitations}

Our approach has some limitations. Like any neural learning method, imitation learning depends heavily on data quality. Initially, we observed degraded performance due to overly fast demonstrations, which compromised precision; slowing the execution significantly improved success rates for tasks such as pallet insertion and stacking. Effective datasets must strike a balance between diversity for generalization and consistency to support efficient learning. Furthermore, computing the relative pose between objects and the end-effector requires consistently accurate estimation of a valid grasping point. Failing to do so limits the agent’s ability to generalize to novel objects, especially after observation filtering by the oracle. Our method relies on an oracle to simplify skill learning, assuming the symbolic solver abstracts information at a sufficiently low level to maintain the Markov property. If not, the learning might not be informed enough, thus leading to the inability to exploit the data for optimal action decision-making. 

Second, our framework supports efficient fine-tuning, enabling fast human-taught robotics. When scaling to more objects of known types, the symbolic domain remains valid; failures arise mainly from neural policies encountering out-of-distribution states, which can be corrected with targeted demonstrations for fine-tuning. If the observation or action space changes due to environmental novelties, the agent must either adapt its planning or learn new controllers. 
In cases where new actions (e.g., screwing), object types (e.g., screwdriver), or predicates (e.g., screwed) are required beyond the original abstraction, performance can usually be restored by extending the graph with a few additional nodes and image pairings. In particular, prior work by Rodriguez et al.~\cite{kr2021rbrg} (Sec.~7.1--7.2) demonstrates strong robustness of the ASP solver to noisy or missing nodes and edges.

Future work includes integrating real human demonstrations into our neuro-symbolic framework and validating it on real systems. While we currently use simulated data, we notably plan to capture expert demonstrations of pallet manipulation and multiple pallets management with a real forklift. Despite the ROS 2 \& Gazebo simulation providing a 1:1 mapping, we want our agent to learn without using the simulation. The main challenge is learning from noisy demonstrations to control the forklift directly. We also aim to demonstrate all the Robosuite scenarios on a real Kinova arm. Additionally, our framework supports future integration of Foundation Models (e.g., CLIP, GPT) to further automate state matching and skill labeling from visual inputs.
\\
\\

\textbf{Acknowledgments}

This work was in part funded by grant N00014-24-1-2024 from the US Office of Naval Research.

%% file: appendix.tex
\appendix                
\section{Appendix}

\begin{figure}[ht]
  \centering
  \begin{minipage}[t]{0.48\textwidth}
    \centering
    \includegraphics[width=\textwidth]{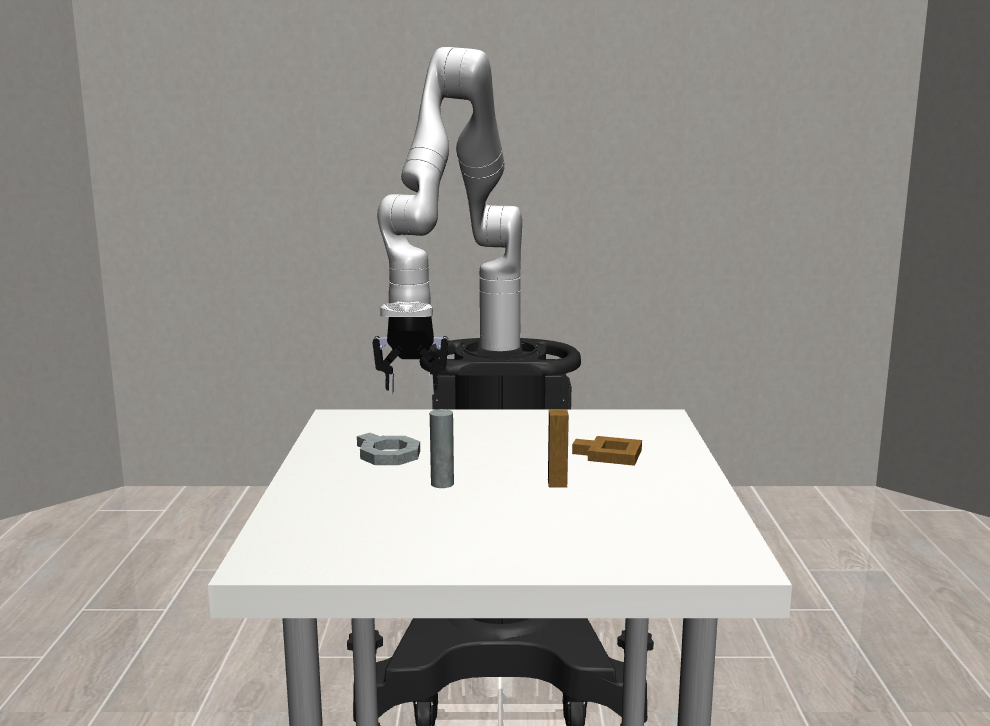}
    \caption{Illustration of the Nut Assembly environment in Robosuite.}
    \label{fig:nutassembly}
  \end{minipage}
  \hfill
  \begin{minipage}[t]{0.48\textwidth}
    \centering
    \includegraphics[width=\textwidth]{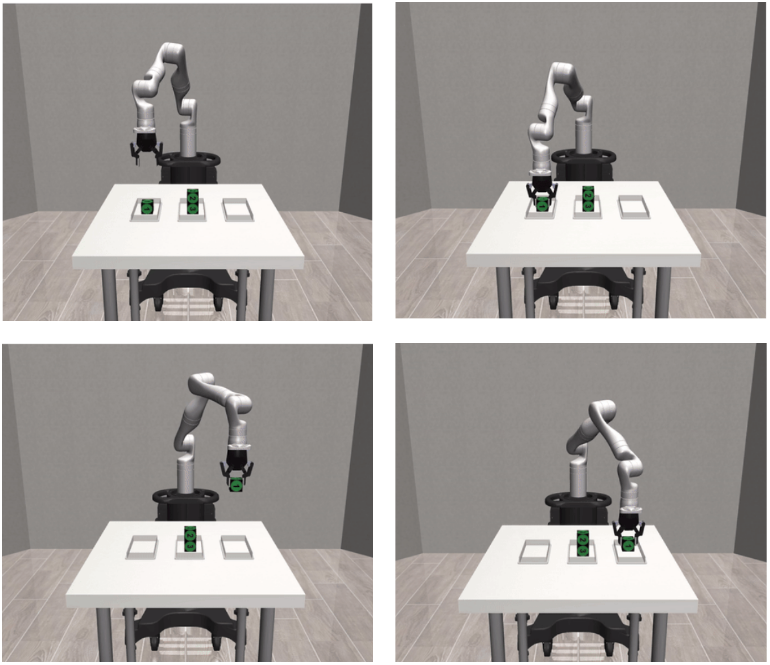}
    \caption{Example of ``MOVE'' skill decomposition into action steps using the \textit{Stacking} and \textit{Towers of Hanoi} demonstrations. From left to right: \textit{reach-pick}, \textit{pick}, \textit{reach-drop}, and \textit{drop}.}
    \label{fig:decomposition}
  \end{minipage}
\end{figure}

\subsection{Symbols Abstraction}
\label{sec:symbols}
\begin{figure}[!b]
    \centering
    \begin{subfigure}[b]{0.43\linewidth}
        \centering
        \includegraphics[width=\linewidth]{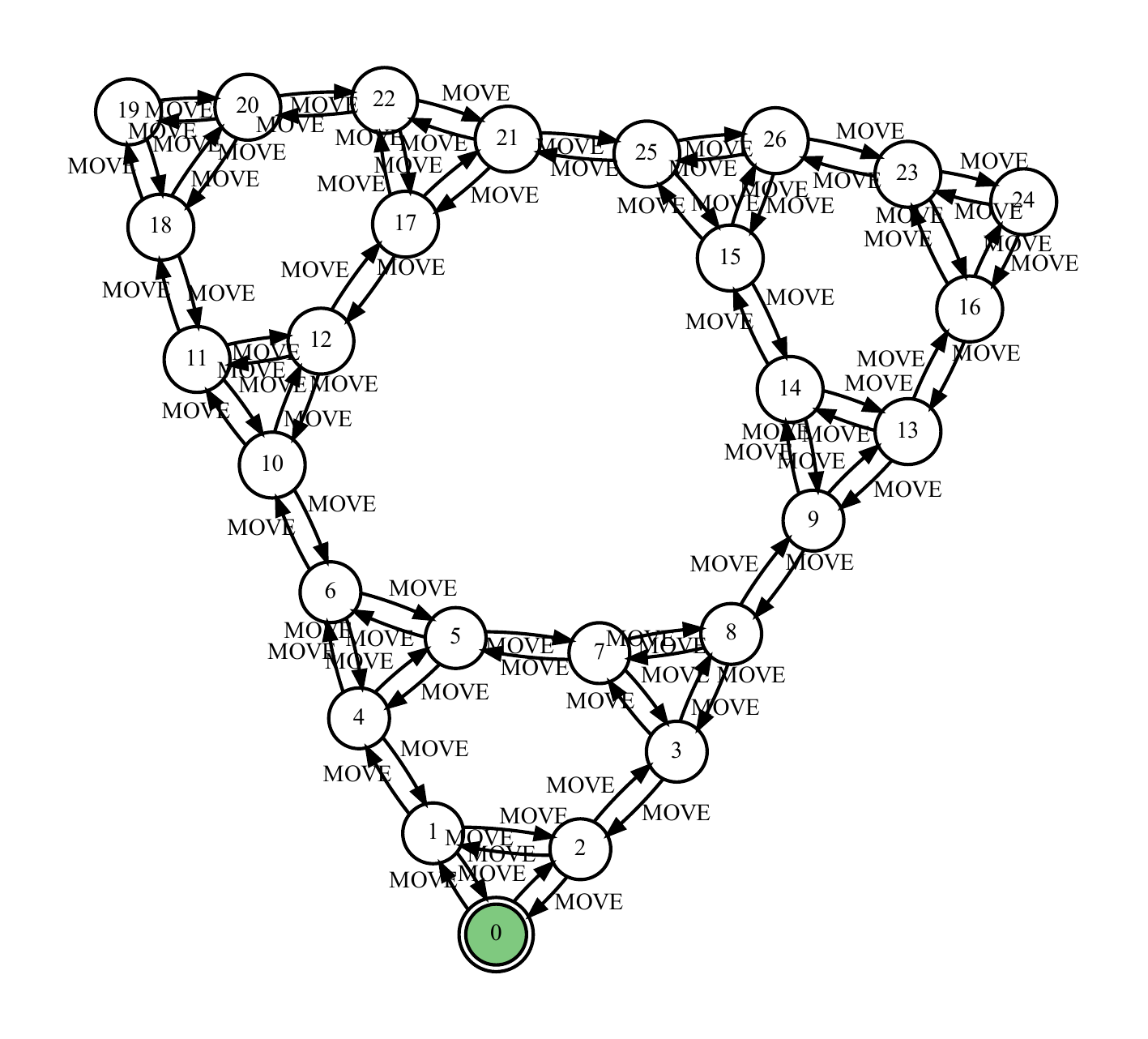}
        \caption{Hanoi Towers Graph}
        \label{fig:fig1}
    \end{subfigure}
    \hfill
    \begin{subfigure}[b]{0.43\linewidth}
        \centering
        \includegraphics[width=\linewidth]{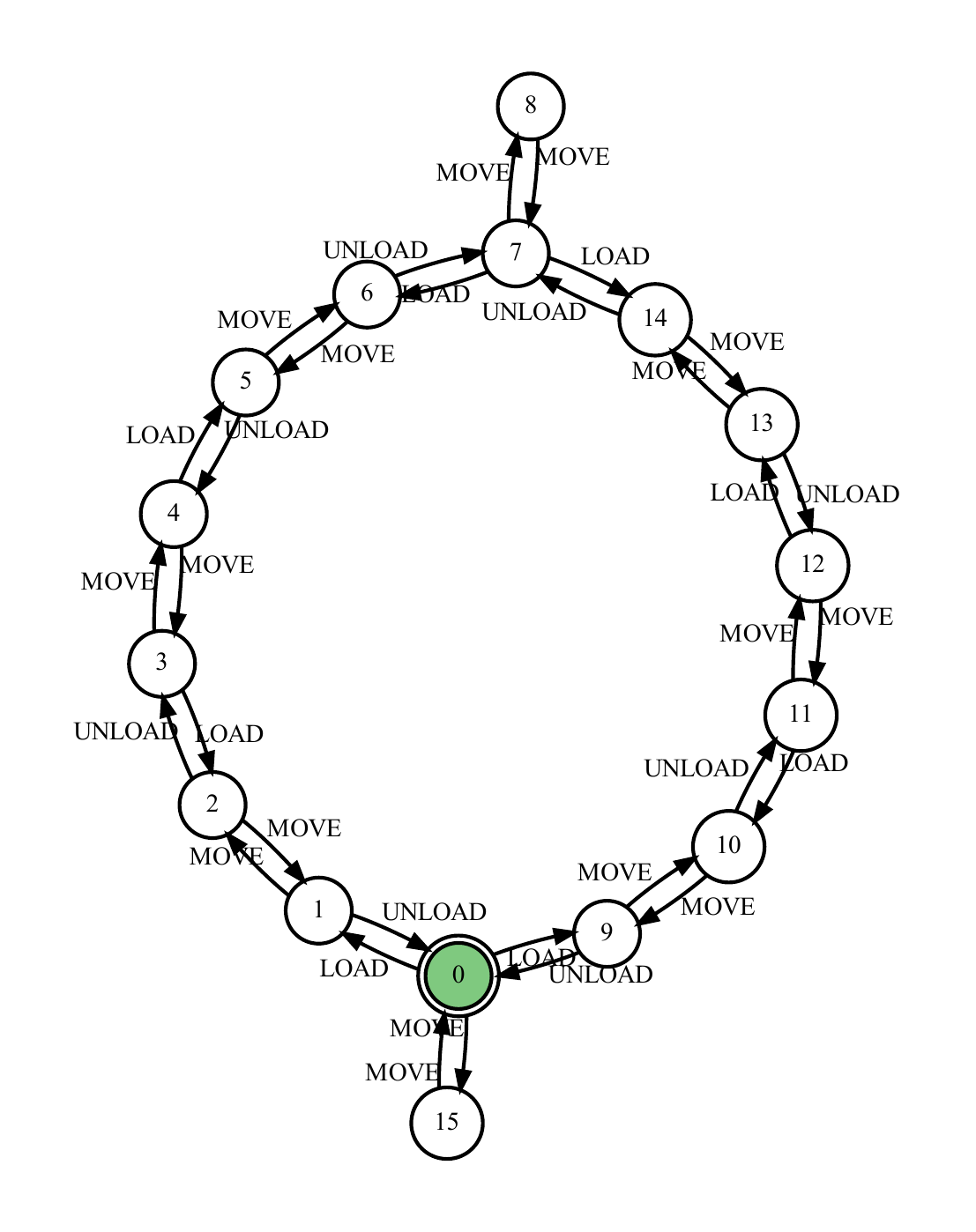}
        \caption{Forklift Multiple Pallets Storage Graph}
        \label{fig:fig2}
    \end{subfigure}

    \vspace{0.5em}

    \begin{subfigure}[b]{0.28\linewidth}
        \centering
        \includegraphics[width=\linewidth]{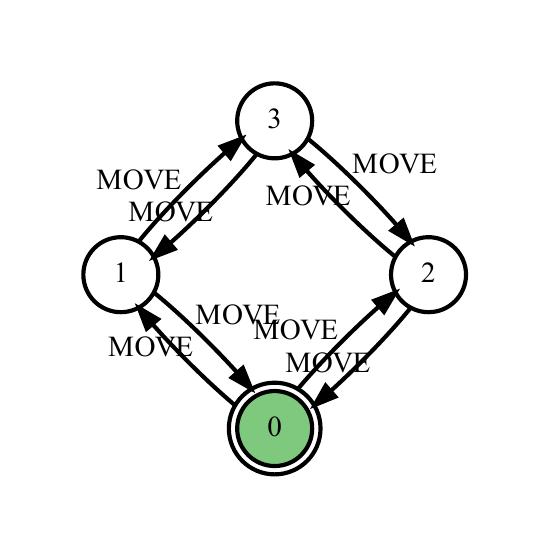}
        \caption{Nut Assembly Graph}
        \label{fig:fig4}
    \end{subfigure}
    \hfill
    \begin{subfigure}[b]{0.60\linewidth}
        \centering
        \includegraphics[width=\linewidth]{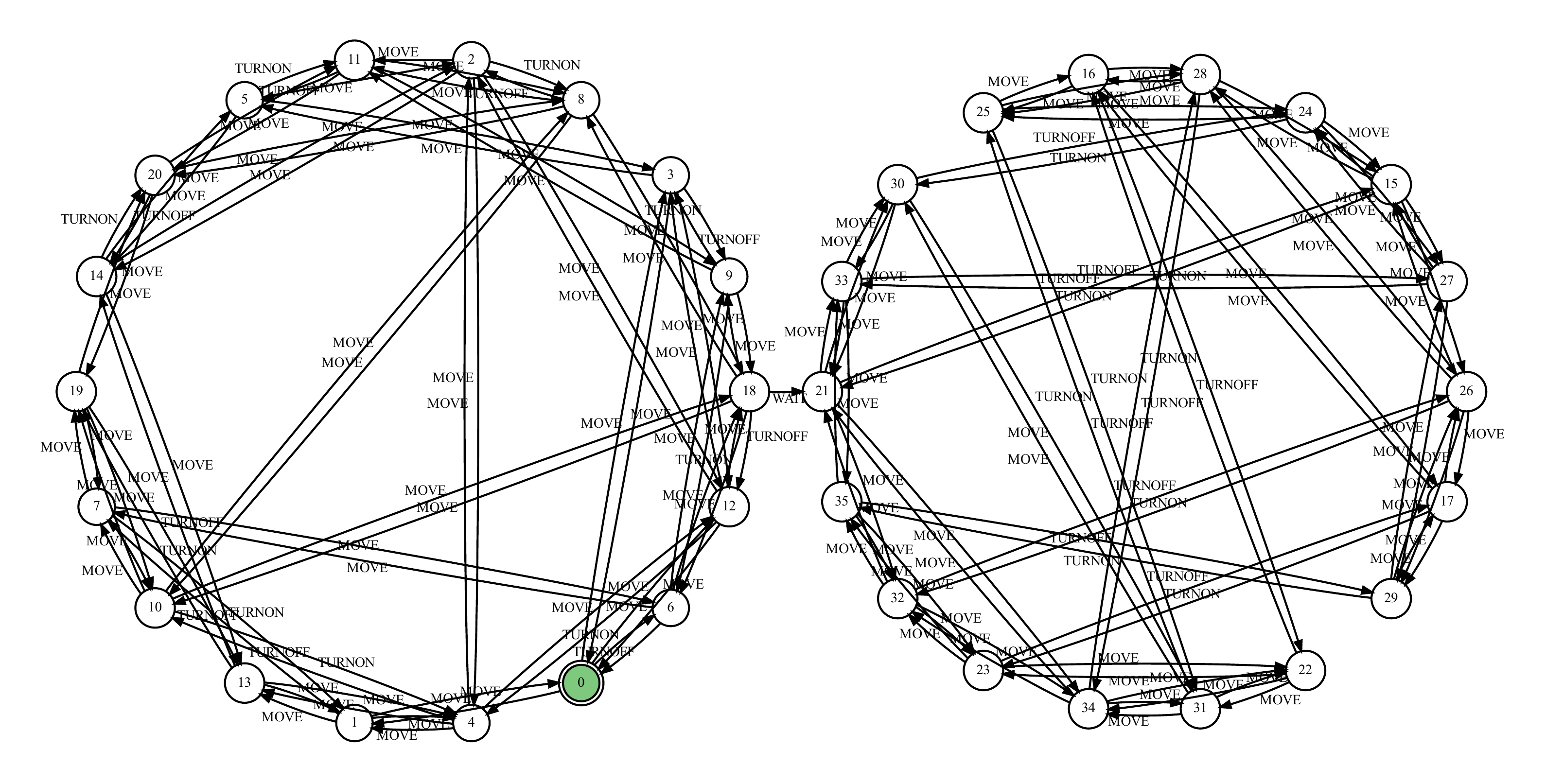}
        \caption{Kitchen Graph (two clusters connected via a unidirectional \texttt{WAIT} \textit{(or COOK)} edge).}
        \label{fig:fig3}
    \end{subfigure}

    \caption{Task graphs for different environments. \textit{Stacking} and \textit{Forklift Load/Unload Pallets} graphs consist only of two nodes and a connecting edge, as they consider one-skill tasks and do not require planning.}
    \label{fig:graph_grid}
\end{figure}

We use a script that automatically transcribes the formatted ASP solver output into PDDL. Note that we have access to both the symbolic operators—i.e., the models of edge transitions in the graph—and the full symbolic description of each node using the entities $\mathcal{E}$ and their relations $\mathcal{F}$. Here are the resulting domains from the graph depicted in Figs~\ref{fig:fig1},~\ref{fig:fig2},~\ref{fig:fig3} and~\ref{fig:fig4}:

\noindent\makebox[\columnwidth]{\rule{0.175\columnwidth}{0.4pt} \textbf{Forklift Multiple Pallets Storage Domain} \rule{0.175\columnwidth}{0.4pt}}

\small  
\footnotesize 

\setlength{\baselineskip}{1.2em}  

  \noindent\underline{MOVE: $a_1(x_1, x_2)$} \\
  \ \ \ \indent Static: $\neg(x_1 = x_2)$ \\
  \ \ \ \indent Pre: $p_2(x_1), p_3(x_2)$ \\
  \ \ \ \indent Eff: $-\!p_2(x_1), -\!p_3(x_2), p_3(x_1), p_2(x_2)$ \\
  \ \ \ \noindent \underline{UNLOAD $a_2(x_1, x_2)$} \\
  \ \ \ \indent Pre: $p_3(x_2), p_4(x_1)$ \\
  \ \ \ \indent Eff: $-\!p_4(x_1), p_1(), p_5(x_1, x_2)$ \\
  \ \ \ \noindent \underline{LOAD: $a_3(x_1, x_2)$} \\
  \ \ \ \indent Pre: $p_1(), p_3(x_2), p_5(x_1, x_2)$ \\
  \ \ \ \indent Eff: $-\!p_1(), -\!p_5(x_1, x_2), p_4(x_1)$ \\
\textbf{Objects:} $o_1, o_2, o_3, o_4$ \textit{(2 pallets x 2 locations)} \\
\noindent\makebox[\columnwidth]{\rule{1\columnwidth}{0.1pt} }
\setlength{\baselineskip}{\normalbaselineskip}  
\normalsize  

\noindent\makebox[\columnwidth]{\rule{0.28\columnwidth}{0.4pt} \textbf{Towers of Hanoi Domain} \rule{0.28\columnwidth}{0.4pt}}

\small  
\footnotesize 

\setlength{\baselineskip}{1.2em}  

  \noindent\underline{MOVE: $a_1(x_1, x_2, x_3)$} \\
  \ \ \ \indent Static: $\neg(x_1 = x_2), \neg(x_1 = x_3), \neg(x_2 = x_3)$ \\
  \ \ \ \indent Pre: $p_1(x_1, x_1), p_1(x_2, x_2), p_1(x_2, x_3), p_2(x_1, x_2)$ \\
  \ \ \ \indent Eff: $-\!p_1(x_1, x_1), -\!p_1(x_2, x_3), p_1(x_2, x_1), p_1(x_3, x_3)$ \\
\textbf{Objects:} $o_1, o_2, o_3, o_4, o_5, o_6$ \textit{(3 disks x 3 pegs)}\\
\noindent\makebox[\columnwidth]{\rule{1\columnwidth}{0.1pt} }
\setlength{\baselineskip}{\normalbaselineskip}  

\normalsize  

\clearpage
\noindent\makebox[\columnwidth]{\rule{0.28\columnwidth}{0.4pt} \textbf{Nut Assembly Domain} \rule{0.28\columnwidth}{0.4pt}}

\small  
\footnotesize 

\setlength{\baselineskip}{1.2em}  

  \noindent\underline{MOVE: $a_1(x_1, x_2, x_3)$} \\
  \ \ \ \indent Pre: $p_1(x_1, x_2)$ \\
  \ \ \ \indent Eff: $-\!p_1(x_1, x_2), p_1(x_3, x_2)$ \\
\textbf{Objects:} $o_1, o_2, o_3, o_4$ \textit{(2 nuts x 2 pegs)}\\
\noindent\makebox[\columnwidth]{\rule{1\columnwidth}{0.1pt} }
\setlength{\baselineskip}{\normalbaselineskip}  

\normalsize  

\noindent\makebox[\columnwidth]{\rule{0.28\columnwidth}{0.4pt} \textbf{Kitchen Domain} \rule{0.28\columnwidth}{0.4pt}}

\small
\footnotesize
\setlength{\baselineskip}{1.2em}

\noindent\underline{MOVE: $a_1(x_1, x_2, x_3)$} \\
  \ \ \ \indent Static: $p_3(x_2, x_1)$ \\
\ \ \ \indent Pre: $p_1(x_2, x_1)$ \\
\ \ \ \indent Eff: $-\!p_1(x_2, x_1),\ p_1(x_2, x_3)$ \\

\noindent\underline{TURNON: $a_2()$} \\
\ \ \ \indent Pre: $\neg p_2$ \\
\ \ \ \indent Eff: $p_2$ \\

\noindent\underline{TURNOFF: $a_3()$} \\
\ \ \ \indent Pre: $p_2$ \\
\ \ \ \indent Eff: $-\!p_2$ \\

\noindent\underline{WAIT: $a_4(x_1)$} \textit{(or COOK)} \\
\ \ \ \indent Pre: $p_2,\ p_1(o_2, o_3),\ p_1(x_1, o_2)$ \\
\ \ \ \indent Eff: $p_4(x_1)$ \\

\textbf{Objects:} $o_1, o_2, o_3, o_4, o_5$ \textit{(e.g., bread, pot, stove, serving-area, table)}\\
\noindent\makebox[\columnwidth]{\rule{1\columnwidth}{0.1pt} }
\setlength{\baselineskip}{\normalbaselineskip}
\normalsize

\subsection{Post-Hoc Interpretation} The output of the ASP Solver does not come with semantics attached, but a human with domain knowledge can interpret such symbols as follows:
\label{sec:symbols}

\noindent\makebox[\columnwidth]{\rule{0.175\columnwidth}{0.4pt} \textbf{Forklift Multiple Pallets Storage Domain} \rule{0.175\columnwidth}{0.4pt}}

\small  
\footnotesize 

\setlength{\baselineskip}{1.2em}  

  \noindent \underline{MOVE: $a_1(x_1, x_2)$} \\
  \ \ \ \indent Pre: \texttt{(free\_location $x_1$), (forklift\_at $x_2$)} \\
  \ \ \ \indent Eff:\texttt{(not (free\_location $x_1$)), (not (forklift\_at $x_2$)), (forklift\_at $x_1$), (free\_location $x_2$)} \\
  \ \ \ \noindent \underline{UNLOAD: $a_2(x_1, x_2)$} \\
  \ \ \ \indent Pre: \texttt{(forklift\_at $x_2$), (loaded\_pallet $x_1$)} \\
  \ \ \ \indent Eff:\texttt{(not (loaded\_pallet $x_1$)), (free\_forklift), (at $x_1$ $x_2$)} \\
  \ \ \ \noindent \underline{LOAD: $a_3(x_1, x_2)$} \\
  \ \ \ \indent Pre: \texttt{(free\_forklift), (forklift\_at $x_2$), (at $x_1$ $x_2$))} \\
  \ \ \ \indent Eff: \texttt{(not (free\_forklift)), (not (at $x_1$ $x_2$)), (loaded\_pallet $x_1$)} \\
\textbf{Objects:} $o_1, o_2, o_3, o_4$ \textit{(2 pallets x 2 locations)}\\
\noindent\makebox[\columnwidth]{\rule{1\columnwidth}{0.1pt} }

\setlength{\baselineskip}{\normalbaselineskip}  

\normalsize  

\noindent\makebox[\columnwidth]{\rule{0.28\columnwidth}{0.4pt} \textbf{Towers of Hanoi Domain} \rule{0.28\columnwidth}{0.4pt}}

\small  
\footnotesize 
\setlength{\baselineskip}{1.2em}  

  \noindent \underline{MOVE: $a_1(x_1, x_2, x_3)$} \\
  \ \ \ \indent Pre: \texttt{(clear $x_1$), (clear $x_2$), (on $x_2$ $x_3$), (greater $x_1$ $x_2$)} \\
  \ \ \ \indent Eff: \texttt{(not (clear  $x_1$)), (not (on $x_2$ $x_3$)), (on $x_2$ $x_1$), (clear $x_3$)} \\

\textbf{Objects:} $o_1, o_2, o_3, o_4, o_5, o_6$ \textit{(3 disks x 3 pegs)}\\
\noindent\makebox[\columnwidth]{\rule{1\columnwidth}{0.1pt} }

\setlength{\baselineskip}{\normalbaselineskip}  

\normalsize  

\clearpage
\noindent\makebox[\columnwidth]{\rule{0.28\columnwidth}{0.4pt} \textbf{Nut Assembly Domain} \rule{0.28\columnwidth}{0.4pt}}

\small  
\footnotesize 

\setlength{\baselineskip}{1.2em}  

\noindent\underline{MOVE: $a_1(x_1, x_2, x_3)$} \\
\ \ \ \indent Pre: \texttt{(on $x_2$ $x_1$)} \\
\ \ \ \indent Eff: \texttt{(not (on $x_2$ $x_1$)), (on $x_2$ $x_3$)} \\
\textbf{Objects:} $o_1, o_2, o_3, o_4$ \textit{(2 nuts x 2 pegs)}\\
\noindent\makebox[\columnwidth]{\rule{1\columnwidth}{0.1pt} }
\setlength{\baselineskip}{\normalbaselineskip}  

\normalsize  

\noindent\makebox[\columnwidth]{\rule{0.28\columnwidth}{0.4pt} \textbf{Kitchen Domain} \rule{0.28\columnwidth}{0.4pt}}

\small
\footnotesize
\setlength{\baselineskip}{1.2em}

\noindent\underline{MOVE: $a_1(x_1, x_2, x_3)$} \\
\ \ \ \indent Pre: \texttt{(on $x_2$ $x_1$), (greater $x_2$ $x_1$)} \\
\ \ \ \indent Eff: \texttt{(not (on $x_2$ $x_1$)), (on $x_2$ $x_3$)} \\

\noindent\underline{TURNON: $a_2()$} \\
\ \ \ \indent Pre: \texttt{(not (stove-on))} \\
\ \ \ \indent Eff: \texttt{(stove-on)} \\

\noindent\underline{TURNOFF: $a_3()$} \\
\ \ \ \indent Pre: \texttt{(stove-on)} \\
\ \ \ \indent Eff: \texttt{(not (stove-on))} \\

\noindent\underline{WAIT: $a_4(x_1)$} \textit{(or COOK)} \\
\ \ \ \indent Pre: \texttt{(stove-on), (on o2 o3), (on $x_1$ o2)} \\
\ \ \ \indent Eff: \texttt{(cooked $x_1$)} \\

\textbf{Objects:} $o_1, o_2, o_3, o_4, o_5$ \textit{(e.g., bread, pot, stove, serving-area, table)}\\
\noindent\makebox[\columnwidth]{\rule{1\columnwidth}{0.1pt} }
\setlength{\baselineskip}{\normalbaselineskip}
\normalsize

\subsection{Skills decomposition}

For \textit{Stacking}, \textit{Nut Assembly} and the 3-disk \textit{Towers of Hanoi} domains, the only action label was MOVE. Following Section.\ref{sec:neurosym_il}\textit{.Intro}, the agent split the policy learning, and broke the MOVE operator down into four action steps: \textit{reach-pick}, \textit{pick}, \textit{reach-drop}, and \textit{drop} all shown in \ref{fig:decomposition}. 

For the forklift \textit{Loading/Unloading Pallet} and \textit{Multiple Pallets Storage}, the agent did not split the given skills into smaller steps.

\subsection{Hardware}

All experiments were conducted on a workstation equipped with an NVIDIA RTX 4090 GPU and 64 GB of RAM. Training and evaluation were performed using PyTorch and MuJoCo/ROS2+Gazebo-based environments. On average, training a single policy from 5 to 30 demonstrations took between 20 minutes to 5 hours, depending on the task complexity, the number of demonstrations and the trajectories average length. Learning the symbolic domain on a CPU only system takes from 0 secondes (\textit{Nut Assembly domain}) to an order of magnitude of a day (\textit{Kitchen domain}). Full evaluation runs (including symbolic domain construction, planning, and policy execution) took approximately 1--3 hours per domain. All experiments were executed using a single GPU and did not require large-scale distributed computing.

\subsection{Experimental Parameters}
\label{app:exp_params}

All policies were trained using the Diffusion Policy framework in a low-dimensional setting for all tasks. No keypoint features were used. The training horizon was set to 16, with 4 observation steps and 8 action steps. Policies were trained using a Transformer-based diffusion model with the following key parameters: 8 layers, 4 attention heads, 256-dimensional embeddings, and causal attention enabled. Dropout was set to 0.0 for embeddings and 0.01 for attention.

The DDPMScheduler was used with 100 training timesteps and a squared cosine beta schedule. The model was trained for 8000 epochs with a batch size of 256, a learning rate of $1 \times 10^{-4}$, and cosine learning rate scheduling with warmup over 1000 steps. The optimizer used Adam with $\beta_1 = 0.9$, $\beta_2 = 0.95$, and a weight decay of $0.1$. Exponential Moving Average (EMA) was applied with an inverse gamma of 1.0 and a cap at 0.9999.

Validation was performed every epoch with a validation split of 2\% and up to 90 episodes used during training. Rollouts and checkpoints were saved every 50 epochs, and model sampling occurred every 5 epochs. 

For dataset handling, we used a padded low-dimensional zarr-based dataset, with appropriate temporal padding before and after each sequence to match training horizon and step requirements.

For demonstration generation, we used a hand-coded automated script and injected Gaussian noise into all action dimensions—except when explicitly set to zero. The noise had a mean equal to half the current target command and a standard deviation of 30\%, promoting trajectory diversity while maintaining task feasibility.